\documentclass[manuscript,screen,sigconf]{acmart}

\setcopyright{none}
\renewcommand\footnotetextcopyrightpermission[1]{}

\usepackage{soul}
\usepackage{url}
\usepackage{hyperref}
\usepackage{graphicx}
\usepackage{amsmath}
\usepackage{algorithm}
\usepackage[noend]{algorithmic}
\usepackage{wrapfig}
\usepackage{xspace,color}
\usepackage{enumitem}
\usepackage{subcaption}
\usepackage{relsize}
\theoremstyle{definition}

\newtheorem{definition}{Definition}
\newtheorem{task}{Task}

\setlist[itemize]{leftmargin=*}
\setlist[enumerate]{leftmargin=*}
\setlist{nosep}

\newcommand{\framework}{\texttt{FRAMM}\xspace}

\begin{document}

\title{FRAMM: \underline{F}air \underline{Ra}nking with \underline{M}issing \underline{M}odalities for Clinical Trial Site Selection}

\author{Brandon Theodorou$^{1}$, Lucas Glass$^{2}$, Cao Xiao$^{3}$ and Jimeng Sun$^{1}$}
\affiliation{
\institution{University of Illinois at Urbana-Champaign$^{1}$}
\country{}}
\affiliation{
\institution{IQVIA$^{2}$}
\country{}}
\affiliation{
\institution{Relativity Inc.$^{3}$}
\country{}}

\begin{abstract}
Despite many efforts to address the disparities, the underrepresentation of gender, racial, and ethnic minorities in clinical trials remains a problem and undermines the efficacy of treatments on minorities. This paper focuses on the trial site selection task and proposes \framework, a deep reinforcement learning framework for fair trial site selection. We focus on addressing two real world challenges that affect fair trial sites selection: the data modalities are often not complete for many potential trial sites, and the site selection needs to simultaneously optimize for both enrollment and diversity since the problem is necessarily a trade-off between the two with the only possible way to increase diversity post-selection being through limiting enrollment via caps. To address the missing data challenge, \framework has a modality encoder with a masked cross-attention mechanism for handling missing data, bypassing data imputation and the need for complete data in training. To handle the need for making efficient trade-offs, \framework uses deep reinforcement learning with a specifically designed reward function that simultaneously optimizes for both enrollment and fairness. 

We evaluate \framework using 4,392 real-world  clinical trials ranging from 2016 to 2021 and show that \framework outperforms the leading baseline in enrollment-only settings while also achieving large gains in diversity. Specifically, it is able to produce a 9\% improvement in diversity with similar enrollment levels over the leading baselines. That improved diversity is further manifested in achieving up to a 14\% increase in Hispanic enrollment, 27\% increase in Black enrollment, and 60\% increase in Asian enrollment compared to selecting sites with an enrollment-only model.
\end{abstract}

\maketitle
\pagestyle{plain}

\section{Introduction}
Clinical trials are the only established process for developing new treatments for diseases. Enrolling sufficient patients from all gender, racial, and ethnic groups are essential for ensuring the treatment's efficacy on all groups. Despite many efforts to address the disparities \cite{sharma2021improving,hughson2016review}, the underrepresentation of minorities in clinical trials remains a problem \cite{knepper2018will,nephew2021accountability}. This consequently undermines the fairness for minorities in obtaining effective treatments. For example, reports shows that African Americans make up $13.4\%$ of the US population, but only $5\%$ of trial participants. Hispanics represent $18.1\%$ of the US population, but less than $1\%$ of trial participants~\cite{yates2020representation}. Also, studies show that due to this enrollment disparity for COVID-19 vaccine trials, the vaccines were estimated to be 30 times less effective for black compared with white patients, and were 250 times less effective for Asian than white patients~\cite{Liu2021-kc}. 

To address the enrollment disparity, existing efforts have included government policy \cite{hwang2022new}, softening eligibility criteria  to make trials more accessible \cite{liu2021evaluating}, and a community engagement-based approach \cite{gray2021diversity}. Recently, deep learning has been introduced to site selection. For example, Doctor2Vec~\cite{biswal2020doctor2vec} proposed to select sites based on predicted patient enrollment. This development brings promise to scalable site selection, but it does not yet have any consideration for diversity.
Thus, there are still the following challenges to be solved:
\begin{enumerate}
    \item \textbf{Missing data modalities across sites}. Different trial sites can have different modalities of features which can be predictive of patient enrollment when seeking to pick sites for future trials. Some of these features such as claims and specialty data can also be missing at different sites. Trial sites with a greater minority population are more likely to have missing data due to insufficient data collection and reporting, so failure to handle this problem only exacerbates the underlying unfairness.
    \item \textbf{The enrollment-diversity trade-off}. While selecting sites only to maximize enrollment can be treated as a prediction task, the addition of diversity adds a unique challenge to the problem. We can not simply constrain fairness by setting minimum percentage thresholds for each group because they would effectively set enrollments caps due to the low minority population selected by enrollment-only models. So, the problem is necessarily a trade-off between enrollment and fairness, and we thus need to optimizing simultaneously for both objectives. \\
\end{enumerate}

\noindent To address these challenges, we propose a deep reinforcement learning framework named \framework , which is enabled by the following technical contributions:
\begin{enumerate}
    \item \textbf{Modality Encoder for Missing Data Handling}. \framework handles missing data by mapping all the diverse modalities into a shared representation space before combining those present into a single site representation. We also introduce a masked cross-attention mechanism as the missing data module using the trial as a query to build a site representation without needing complete site features.
    \item \textbf{Deep Reinforcement Learning for Efficient Trade-offs}.  \framework is equipped with a deep reinforcement learning setup with a specifically built reward function that simultaneously optimizes for both enrollment and fairness metrics. It also has a deep Q-Value network which approximates the contribution of each individual site to the corresponding reward given their site features. 
\end{enumerate}

We evaluate \framework using 4,392 real-world  clinical trials ranging from 2016 to 2021 from a large clinical trial company. We show that \framework outperforms the leading baseline in enrollment-only settings while also achieving large gains in diversity. Specifically, it is able to produce a 9\% improvement in diversity with similar enrollment levels over the leading baselines. That improved diversity is further manifested  in achieving up to a 14\% increase in Hispanic enrollment, 27\% increase in Black enrollment, and 60\% increase in Asian enrollment compared to selecting sites with an enrollment-only model.

\section{Related work}

\paragraph{Machine Learning for Clinical Trials}
There have been a number of recent applications which look at using machine learning to optimize clinical trial operations. These include matching patients to trials that they are eligible for \cite{gao2020compose,zhang2020deepenroll}, searching for similar trials \cite{wang2022trial2vec}, and predicting trial outcomes \cite{fu2022hint}. There have even been some works seeking to predict site enrollments to help select trial sites \cite{biswal2020doctor2vec,gligorijevic2019optimizing}. However, none of the existing works were designed to optimize enrollment diversity. 

\paragraph{Missing Data Handling}
Existing missing data handling mainly relies on data imputation. The existing methods typically reconstruct embeddings of missing modalities based on the embeddings of present ones as in \cite{ma2021smil,Tran_2017_CVPR,lau2019unified}. This requires some complete data points and pre-training of the imputation model. Other approaches include modality dropout during training \cite{parthasarathy2020training}, or learning multiple conditional distributions for different combinations of present modalities \cite{ma2021maximum}.
Our framework can entirely bypass data imputation and the need for complete data in model training via a cross attention mechanism.

\section{Problem Formulation}


\begin{definition}[\textbf{Clinical Trial Site Features}]
For each clinical trial, the trial is represented by a vector $\mathbf{t} \in \mathbb{R}^{n_t}$ containing features for disease information, trial logistics, and eligibility information. Its sites then contain the following data modalities: (1) \textbf{Static information} $\mathrm{s}_i \in \mathbb{R}^{n_s}$ is a vector representing a site's primary clinician's gender, profession type, primary specialty, patient demographic distributions, and geo coordinates of the site; (2) \textbf{Diagnosis history}  $\mathrm{D}_i \in \mathbb{R}^{n_c \times n_d}$ is a sequence of $n_c$ one-hot vectors representing the ICD-10 codes (out of $n_d$  options) of the diagnoses of the most recent patients for a given site; (3) \textbf{Medication history} $\mathrm{P}_i \in \mathbb{R}^{n_c \times n_p}$ is a sequence of $n_c$ one-hot vectors representing for the Uniform System of Classification ontology level 2 \cite{USC} codes (out of $n_p$ options) of the most recent prescriptions at a given site; and (4) \textbf{Enrollment history} $\mathrm{E}_i \in \mathbb{R}^{n_h \times (n_t')}$ is a sequence of $n_h$ trial representations (omitting the inclusion/exclusion criteria due to dimensionality/memory concerns) and enrollment numbers for the most recent trials of a site.

In addition, trial sites also have an input feature mask $\mathbf{m}_i \in \mathbb{R}^{4}$ represented as a 4-dimensional binary vector signifying whether each feature modality is present or missing. Finally, trial sites are labeled using their enrollment value for the trial $e_i \in \mathbb{R}$ and their 6-dimensional vector $\mathbf{r}_i \in \mathbb{R}^{6}$ representing their racial distribution. For example, if Site 4 enrolled 95 participants and has a racial makeup of 47\% White, 23\% Hispanic, 15\% Black, 10\% Asian, 4\% Mixed, and  1\% Others, we have $e_4 = 95$ and $\mathbf{r}_4 = [47, 23, 15, 10, 4, 1]$. 

    

So, each trial $\mathbf{t}$ is scored against $M$ trial sites, which are each represented as $S_i = \big((\mathbf{s}_i, \mathbf{D}_i, \mathbf{P}_i, \mathbf{E}_i), (\mathbf{m}_i), (e_i, \mathbf{r}_i)\big)$ where $(\mathbf{s}_i, \mathbf{D}_i, \mathbf{P}_i, \mathbf{E}_i)$ are feature modalities,  $\mathbf{m}_i$ is the feature mask, and $(e_i, \mathbf{r}_i)$ the labels of enrollment value $e_i$ and racial distribution $\mathbf{r}_i$. 
\end{definition}

\begin{task}[\textbf{Clinical Trial Site Selection}]
For an input trial, the task is to select $K$ sites from its M choices to maximize overall diversity and enrollment. Mathematically, given a trial $\mathbf{t}$ and its $M$ prospective sites $[S_1, S_2, \cdots, S_M]$, the task becomes to select $K$ of those sites based on their present input features to maximize the received reward $R(\mathcal{R})$.
In this paper, this is achieved by ranking the $M$ sites as $\mathcal{R}$, where $\mathcal{R}_j$ is the $j$-th site in the ranking, and selecting the $K$ highest ranked sites.  The ranking $\mathcal{R}$ then produces two additional outputs:
\begin{itemize}
    \item $\mathrm{\tilde{e}} \in \mathbb{R}^M$, a vector of the enrollment values ($e_i$'s) of each site in the ranked order;
    \item $\mathrm{\tilde{R}} \in \mathbb{R}^{M \times 6}$ which is the 6 racial distribution values ($\mathrm{r}_i$'s) for each of the sites in the ranked order.
\end{itemize}
\end{task}

\noindent The notations are summarized in Table \ref{fig:NotationTable} for reference.

\begin{table}
\centering
\caption{Table of Notations}
\resizebox{0.95\columnwidth}{!}{
\begin{tabular}{c|l} \toprule
Notation   & Description \\ \midrule
$M\in\mathbb{N}$          & The number of site options for a trial \\
$K\in\mathbb{N}$          & The number of sites to select from the $M$ \\
$\lambda\in\mathbb{R}$  & The relative weighting of utility and fairness \\ \hline
$\mathbf{t}\in\mathbb{R}^{n_t}$ & A vector representation of a clinical trial \\ \hline
$S_i$       & The $i$-th site option (of $M$) for the trial \\ \hline
$\mathbf{s}_i\in\mathbb{R}^{n_s}$  & The $i$-th site's static features modality \\
$\mathbf{D}_i\in\mathbb{R}^{n_c \times n_d}$  & The $i$-th site's diagnosis history modality \\
$\mathbf{P}_i\in\mathbb{R}^{n_c \times n_p}$  & The $i$-th site's prescription history modality \\
$\mathbf{E}_i\in\mathbb{R}^{n_h \times (n_t' + 1)}$  & The $i$-th site's enrollment history modality \\ \hline
$\mathbf{m}_i\in\mathbb{R}^4$  & The $i$-th site's modality presences \\ \hline
$e_i\in\mathbb{N}$  & The $i$-th site's enrollment for the trial \\
$\mathbf{r}_i\in\mathbb{R}^6$  & The $i$-th site's racial distribution vector \\ \hline
$R$  & The reward function for a selection of sites \\ \hline
$\mathcal{R}$  & An ordered ranking of the $M$ sites \\
$\mathbf{e}\in\mathbb{R}^M$  & The site enrollment numbers ordered by $\mathcal{R}$ \\
$\mathbf{R}\in\mathbb{R}^{Mx6}$  & The racial distributions ordered by $\mathcal{R}$ \\ \bottomrule
\end{tabular}}
\label{fig:NotationTable}
\end{table}

\begin{figure*}
    \centering
    \includegraphics[scale=0.59]{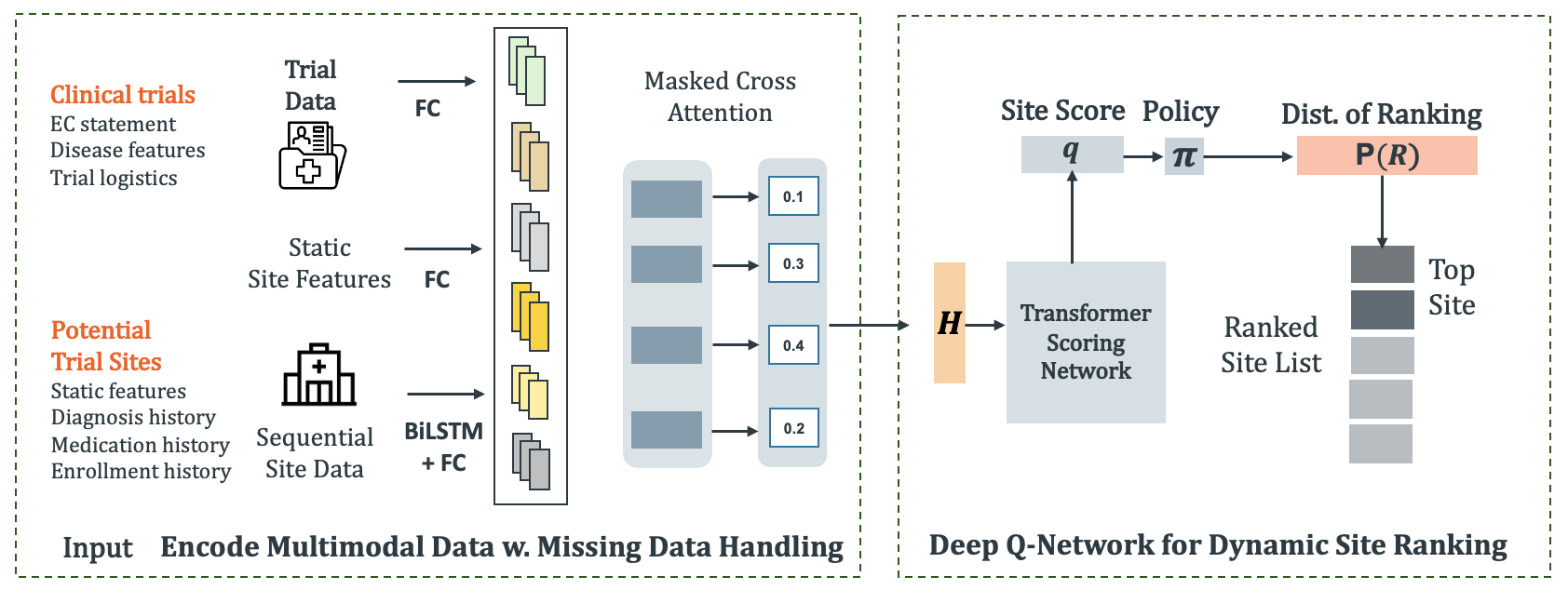}
    \caption{A visualization of the \framework framework. \framework uses multi-modal site features and the trial representation to generate scores for, rank, and select a subset of prospective trial sites. The pipeline used to do so consists of modality encoders, a missing data handling mechanism, a scoring network, and a reinforcement learning-based ranking approach.  }
    \label{fig:FRAMM_Architecture}
\end{figure*}

\section{FRAMM}
Our \framework framework ranks and selects potential trial sites with the following two modules: 

\noindent
(1) \textbf{the missing data modality encoders} that generate trial-site representations while handling the fact that some of the input site modalities embeddings may be missing.

\noindent
(2) \textbf{the scoring and ranking network} that maps each trial-site representation to a single score approximating the value of a site to a particular clinical trial before converting the scores into a probability distribution over different rankings according to a learned policy.

\subsection*{(I) Modality Encoding with Missing Data Handling}

For each of the $M$ sites (where subscript $i$ throughout refers to the $i$-th site of the $M$), the modality encoders first embed each available modalities and trial representation into a shared representation space, $\mathbb{R}^{n_{\text{emb}}}$ (where we use $n_{\text{emb}} = 128$). 

For the three sequential modalities, diagnosis $\mathbf{D}_i$, prescription $\mathbf{P}_i$, and enrollment history $\mathbf{E}_i$, we embed the inputs using $f$, parameterized for the different settings by $d$, $p$, and $e$, respectively such that they map $\mathbf{d^{e}}_i = f_d(\mathbf{D}_i)$, $\mathbf{p^{e}}_i = f_p(\mathbf{P}_i)$, and $\mathbf{e^{e}}_i = f_e(\mathbf{E}_i)$. In our experiments, we implement $f$ with a bidirectional LSTM followed by a fully connected neural network with ReLU activation function after the first linear layer as in Eq.~\eqref{eq:bilstm}.
\begin{equation}\label{eq:bilstm}
    f(x) = \max(0,\text{biLSTM}(x)\mathbf{W} + \mathbf{b})\mathbf{V} + \mathbf{c}
\end{equation}
where the LSTM has hidden dimension 128, $\mathbf{W},\mathbf{V} \in \mathbb{R}^{128 \times 128}$, and $\mathbf{b},\mathbf{c} \in \mathbb{R}^{128}$.

We then embed the static site modality and trial representation using $g_s$ and $g_t$ respectively such that they map $\mathbf{s^{e}}_i = g_s(\mathbf{s}_i)$ and $\mathbf{t^{e}}_i = g_t(\mathbf{t})$. We implement $g$ as a  fully-connected neural network with ReLU activation function between layers as in Eq.~\eqref{eq:fc},
\begin{equation}\label{eq:fc}
    g(x) = \max(0,x\mathbf{W} + \mathbf{b})\mathbf{V} + \mathbf{c}
\end{equation}
where $\mathbf{W} \in \mathbb{R}^{\text{dim}(x) \times 128}$, $\mathbf{V} \in \mathbb{R}^{128 \times 128}$, and $\mathbf{b},\mathbf{c} \in \mathbb{R}^{128}$. So, for each site we have access to 4 possibly missing modality embeddings $\mathbf{s^e}$, $\mathbf{d^e}$, $\mathbf{p^e}$, and $\mathbf{e^e}$ and a trial embedding $\mathbf{t^e}$ all within the shared representation space $\mathbb{R}^{n_{\text{emb}}}$. These five embeddings are then fed into the missing data mechanism within the module.

\paragraph{Missing Data Handling}
To handle missing data modalities, existing strategies, including Modality Dropout (MD)~\cite{parthasarathy2020training}, Unified Representation Network (URN)~\cite{lau2019unified}, and Cascaded Residual Autoencoder(CRA)~\cite{Tran_2017_CVPR}  either do not directly model missing data or require pre-training. To avoid these issues, we propose \textbf{Masked Cross-Attention (MCAT)} which uses the trial embedding as the query for a masked multi-head cross-attention mechanism where the site modality embeddings serve as both keys and values, and the site feature masks dictate whether a given modality can be attended to. The output of this mechanism is the intermediate site embedding, $\mathbf{h}'_i$ which is then concatenated back with the trial embedding to become the trial-site representation $\mathbf{h}_i$.
More formally, let the missing data handling procedure be $miss$ such that $\mathbf{h}_i = miss([\mathbf{s^{e}}_i, \mathbf{d^{e}}_i, \mathbf{p^{e}}_i, \mathbf{e^{e}}_i, \mathbf{t^{e}}_i])$.
For a given site, this MCAT approach arrives at $\mathbf{h}_i$ by Eq.~\eqref{eq:mcat},
\begin{align}\label{eq:mcat}
    &\mathbf{h}'_i = \text{att}(\mathbf{t^{e}}_i, [\mathbf{s^{e}}_i, \mathbf{d^{e}}_i, \mathbf{p^{e}}_i, \mathbf{e^{e}}_i], [\mathbf{s^{e}}_i, \mathbf{d^{e}}_i, \mathbf{p^{e}}_i, \mathbf{e^{e}}_i]) \nonumber \\
    &\textrm{att}(Q,K,V) = \text{concat}(\text{head}_1, \cdots, \textrm{head}_4)\mathbf{W^O} \nonumber \\
    &\textrm{head}_j = \sigma\left(\frac{(Q \mathbf{W}^\mathbf{Q}_j) (K \mathbf{W}^\mathbf{K}_j)^T}{\sqrt{n_{emb}/n_{\text{head}}}} + \mathbf{m'}_i\right)(V \mathbf{W}^\mathbf{V}_j) \\
    & m'_{i k} =\begin{cases} 
        0 &m_{i k} = 1 \nonumber \\
        -\infty &m_{i k} = 0
    \end{cases}\\
    &\mathbf{h}_i = \text{concat}(\mathbf{h}'_i, \mathbf{t^{e}}_i) \nonumber
\end{align}
where $\sigma$ is the softmax function, $\mathbf{W}^\mathbf{Q}_j, \mathbf{W}^\mathbf{K}_j, \mathbf{W}^\mathbf{V}_j, \mathbf{W^O} \in \mathbb{R}^{128 \times 128}$, and $\mathbf{m'}_i \in \mathbb{R}^4$ is a conversion of the $i$-th site's binary mask vector into the form required for masking in attention. That conversion converts a mask value of $0$ (signifying a missing modality) to $-\infty$ to prevent any remaining attention weight after a Softmax function over the sum of the values and masks, and it converts a value of $1$ (signifying a present modality) to $0$ to not interfere with the Softmax calculation.

\subsection*{(II) Deep Q-Network and Ranking Policy}
The second module takes the $M$ trial-site representations, now without any missingness, and uses them to rank the sites. To do so, it first maps the representations $\mathbf{H} = [\mathbf{h}_1 \cdots \mathbf{h}_M]$ to $M$ real-valued scores $\mathbf{q} \in \mathbb{R}^M$ which approximate the value each site will provide towards the final reward obtained by selecting them for the current trial. The $\mathbf{q}$ here serves an analogous (although not identical) role to a Q-Value in reinforcement learning as in Eq.~\eqref{eq:q},
\begin{equation}\label{eq:q}
    Q(s,a) = R(s,a) + \gamma \max_{a'} Q(s',a')
\end{equation}
This mapping is denoted generally by the $score$ function in Eq.~\eqref{eq:score},
\begin{equation}\label{eq:score}
    \mathbf{q} = score(\mathbf{H})
\end{equation} 

\paragraph{Reward Function}
The reward function used in this paper and which the Q-Network learns to approximate consists of two components for enrollment utility and fairness objectives and is defined by Eq.~\eqref{eq:rewardfunction},
\begin{equation}\label{eq:rewardfunction}
    R(\mathcal{R}) = U(\mathcal{R}) + \lambda F(\mathcal{R})
\end{equation}
where $\lambda$ is a hyperparameter determining the relative weighting of the two reward components. This relative weighting forms a trade-off such that multiple $\lambda$ values can be utilized and the desired point along the trade-off curve can be selected. We chose $\lambda$ as 0, 0.5, 1, 2, 4, or 8 in our experiments. 

The utility component is defined by Eq.~\eqref{eq:fair},
\begin{equation}\label{eq:fair}
    U(\mathcal{R}) = \frac{ \sum_{j=1}^{K}\tilde{e}_j - \sum_{j=K+1}^{M}\tilde{e}_j}{\sum_{j=1}^{M}\tilde{e}_j}
\end{equation}
which is the enrollment difference between the chosen sites (top-$K$ sites) and those not (the remaining $M-K$ sites), normalized for the total enrollment numbers.

The fairness component is the entropy of the racial distribution of the total patient population enrolled by the $K$ chosen sites, formally defined by Eq.~\eqref{eq:entropy},
\begin{equation}\label{eq:entropy}
    F(\mathcal{R}) = H\left(\frac{\sum_{j=1}^{K}\tilde{e}_j \mathbf{\tilde{r}}_j}{\sum_{j=1}^{K}\tilde{e}_j}\right)
\end{equation}
where both $\tilde{e}_j$ (the enrollment values in the ranked order) and the bottom sum are scalars applied to the entire 6-dimensional vector $\mathbf{\tilde{r}}_j$, the top sum operates element-wise over those vectors, and $H$ is the standard definition of entropy where for a probability distribution $\mathbf{n}$ (such as the one outputted by our sum), $H(\mathbf{n}) = - \sum_k n_k \log n_k$.

\paragraph{Q-Network Architecture}
We implement this mapping using a transformer encoder layer with a fully-connected head. We postulate that transformer layers and their ability for each site's score to be affected by the other sites available can be especially valuable given the dynamic nature of $F$ in which the fairness component of our reward evaluates all of the selected sites together rather than acting as a sum of functions of each individual site as with the utility component.
The generation of the scores is formally given by Eq.~\eqref{eq:tr},
\begin{align}\label{eq:tr}
    &\mathbf{H}^{(0)} = \mathbf{H} \nonumber \\
    &\mathbf{H'}^{(i)} = \mathrm{LN}(\mathbf{H}^{(i-1)} + \text{att}(\mathbf{H}^{(i-1)}, \mathbf{H}^{(i-1)}, \mathbf{H}^{(i-1)})) \\
    &\mathbf{H}^{(i)} = \mathrm{LN}(\mathbf{H'}^{(i)} + (\max(0,\mathbf{H'}^{(i)} \mathbf{W}^{(i)} + \mathbf{b}^{(i)}) \mathbf{V}^{(i)} + \mathbf{c}^{(i)})) \nonumber \\
    &\mathbf{q} = \max(0,\mathbf{H}^{(n_l)} \mathbf{W}^{f} + \mathbf{b}^{f}) \mathbf{V}^{f} + \mathbf{c}^{f} \nonumber
\end{align}
where $i \in \{ 1, \cdots, n_l \}$, LN denotes Layer Normalization, $\mathbf{W}^{(i)}, \mathbf{V}^{(i)} \in \mathbb{R}^{128 \times 128}$, $\mathbf{b}^{(i)}, \mathbf{c}^{(i)} \in \mathbb{R}^{128}$, $\mathbf{W}^f \in \mathbb{R}^{128 \times 64}$, $\mathbf{b}^f \in \mathbb{R}^{64}$, $\mathbf{V}^f \in \mathbb{R}^{64 \times 1}$, and $\mathbf{c}^f \in \mathbb{R}$.

\paragraph{Ranking and Policy Learning}
These ``Q-Value'' scores are then fed into the final portion of our framework where they are used to generate rankings and their probabilities using a non-deterministic policy $\pi$. While simply selecting the top $K$ scores represents the best site selection as currently approximated by the network (and is used for testing), it prevents exploration of the ranking space during training. Instead, we define our stochastic policy by Eq.~\eqref{eq:stoc},
\begin{equation}\label{eq:stoc}
    \pi(\mathcal{R}) = \sum_{\mathcal{R'} \in \phi(\mathcal{R})}\prod_{j=1}^K \frac{\exp(q(\mathcal{R'}_j))}{\sum_{k = j}^M \exp(q(\mathcal{R'}_k))}
\end{equation}
where $\phi$ maps to the set of other rankings which are the same as $\mathcal{R}$ as  except for permuting the first $K$ elements, and the $q$ function returns the score given by the Q-Value network for the given site. Our policy's probability represents the odds of a given top-$K$ combination (where order does not matter) achieved by sampling $K$ sites without replacement according to the softmax probabilities defined by their scores. However, in practice, this quickly becomes impractical for any sizable $K$ as the permutation space grows factorially. Instead, we use an unbiased estimate of the probability of a top-$K$ combination calculated for a given ranking by randomly permuting the order of the first $K$ elements (to remove the bias of higher probability rankings in the permutation space being drawn more often), calculating the product in the policy's equation, and scaling by $K!$. In this way, we are able to estimate according to the expectation in Eq.~\eqref{eq:exp},
\begin{align}\label{eq:exp}
    \pi(\mathcal{R}) &= \sum_{\mathcal{R'} \in \phi(\mathcal{R})}\prod_{j=1}^K \frac{\exp(q(\mathcal{R'}_j))}{\sum_{k = j}^M \exp(q(\mathcal{R'}_k))} \nonumber \\
        &= K! \mathop{\mathlarger{\mathlarger{\mathbb{E}}}}_{\mathcal{R'} \in \phi(\mathcal{R})}\prod_{j=1}^K \frac{\exp(q(\mathcal{R'}_j))}{\sum_{k = j}^M \exp(q(\mathcal{R'}_k))}
\end{align}

Given our architecture for obtaining scores for each site, and our method of sampling and obtaining probability estimates of rankings given those scores, all that is left is to handle optimization and policy learning. The overall goal is to maximize the expected reward in Eq.~\eqref{eq:reward},
\begin{equation}\label{eq:reward}
    \mathbb{E}_{\mathcal{R}\sim\pi}[R(\mathcal{R})]
\end{equation}

We implement this using  REINFORCE~\cite{williams1992simple}. It is a common policy gradient algorithm which directly optimizes the expected reward as calculated through Monte Carlo sampling weighted by the log-likelihoods of rankings. Using this or any other method policy gradient algorithm, we are able to use backpropagation back through our policy, Q-Network, and Modality Encoders to train our framework.


\section{Evaluation}

\subsection{Experimental Setup}

\paragraph{Datasets} We use real-world clinical trials and claims data in evaluation. The clinical trial database contains 33,323 sites matched with 4,392 trials. We first build the site pool by constructing input features. We then create a separate dataset for each values of $M$ that we use. We match each trial to $M$ sites, using the top $M$ sites (determined by enrollment) in the database if there are enough and otherwise completing the set of $M$ by randomly selecting sites from the overall pool and assigning an enrollment of $0$ for the trial. Finally, we add in the missing data aspect by creating 10 versions of each trial, and for each site in each trial randomly creating a mask which dictates whether a given site modality is present for that data point (where each modality has an 80\% chance of being present). After this augmentation, we are left with our final dataset for the given $M$ value.

We randomly split these datasets by trial into training and test datasets with an 80-20 ratio and then further split off 10\% of the training set into a validation set. Using these datasets, we train our models within the PyTorch framework \cite{PyTorch} for 35 epochs at a 0.00001 learning rate and using the Adam optimizer. We save the model which best performs on the validation set as determined by our reward function and evaluate it using the test set.

\paragraph{Baselines} We consider the following baseline models
\begin{enumerate}
    \item{\textbf{Doctor2Vec}~\cite{biswal2020doctor2vec} is the current state of the art in enrollment-only trial site selection. It constructs a memory network doctor representation based on static features and patient visits. That representation is then queried by a specific trial representation and fed into a downstream network to predict the doctor's enrollment count for the trial. Note that Doctor2Vec does not handle missing data and so is trained on the smaller, full-data version of our dataset before each trial was repeated 10 times.}
    \item{\textbf{Random} selects $K$ sites at random from the available $M$.}
    \item{\textbf{One-Sides Policy Gradient (PGOS)}~\cite{singh2019policy} is a fairness baseline that replaces the fairness function $F$ with a one-sided loss function which ensures through regularization that groups are not underrepresented within rankings. This baseline represents the typical approach of constraining or regularizing fairness rather than explicitly optimizing diversity. Note that when $\lambda = 0$, $F$ does not contribute to the overall reward, and this is identical to our standard framework. So, it is omitted from any enrollment-only results.}
\end{enumerate}

\paragraph{Ablation Models} To demonstrate \framework's effectiveness at handling missing data and its ability to use missingness as a data augmentation technique to combat low data settings such as ours (the original, complete dataset has less than 5,000 trials), we add two ablation models trained on the same smaller, full-data dataset as Doctor2Vec. The first is the full \framework model trained on this smaller dataset, and the second removes the missing data mechanism and replaces it with a fully connected layer. We call these models `FRAMM No Missing' and `FC No Missing' respectively.






\paragraph{Metrics} We consider both enrollment and diversity metrics.

\noindent
For enrollment, we compared the size of each model's enrolled cohort with the ground truth maximal enrollment via a pair of metrics. First, we use relative error calculated by
\begin{equation}
    \text{Relative Error} = \frac{\text{Max Enrollment} - \text{Model Enrollment}}{\text{Max Enrollment}}
\end{equation}
where max enrollment is the total enrollment from the top $K$ sites after the trial completion (a theoretical ceiling), and model enrollment is the total enrollment from the $K$ sites selected by the model. We also report the standard ranking metric normalized Discounted Cumulative Gain (nDCG), defined
\begin{equation}
    \text{nDCG} = \sum_{j=1}^{K} \frac{2^{m_j} - 1}{\log_2(j+1)} \big{/} \sum_{j=1}^{K} \frac{2^{o_j} - 1}{\log_2(j+1)}
\end{equation}
where $m$ is the model ranking enrollment list, and $o$ is the optimal ranking enrollment list. For example, if we have four sites A, B, C, and D with enrollment values of 5, 10, 8, and 7 respectively, and our model ranked them B $\rightarrow$ C $\rightarrow$ A $\rightarrow$ D, then we would have $m = [10, 8, 5, 7]$ and $o = [10, 8, 7, 5]$.

\noindent To measure diversity, we use the entropy of the overall racial distribution of the final enrolled population. This is defined in the same way as above within our reward function, $F$, by
\begin{equation}
    H(\mathbf{p}) = - \sum_{k=1}^6 p_k \log p_k
\end{equation}
where $\mathbf{p}$ is the vector of the proportions of each racial group within the overall enrolled population and so $p_k$ is the percentage of a given group.


\subsection{Results}

We design experiments to answer the following questions.
\begin{enumerate}
    \item Is \framework effective at enrolling large patients populations in enrollment-only settings?
    \item Can \framework make efficient trade-offs between enrollment and diversity to achieve high levels of both?
    \item Does \framework improve diversity compared to enrollment-only and post-hoc constrained models?
\end{enumerate}

\begin{table}
\caption{Enrollment-Only Performance}
\begin{tabular}{l|cc}
\toprule
&Relative Error ($\downarrow$)    &nDCG ($\uparrow$) \\ \midrule
Random  &0.621 $\pm$ 0.019     &0.320 $\pm$ 0.017 \\
Doctor2Vec &0.525 $\pm$ 0.021     &0.402 $\pm$ 0.018 \\ 
FRAMM No Missing   &0.572 $\pm$ 0.020     &0.359 $\pm$ 0.018 \\
FC No Missing    &0.566 $\pm$ 0.020  &0.363 $\pm$ 0.017 \\\hline
FRAMM    &\textbf{0.512} $\mathbf{\pm}$ \textbf{0.020}     &\textbf{0.409} $\mathbf{\pm}$ \textbf{0.018}
\\ \bottomrule
\end{tabular}
\label{tab:Enrollment}
\end{table}

\subsection*{Q1. Enrolling Large Patient Populations} 
We first evaluate each model in enrollment-only settings (with $\lambda = 0$ for \framework variants) to examine their ability to select sites with only enrollment in mind. We display both of our enrollment metrics for the $M=20$, $K=5$, $\lambda=0$ setting for each compared model in Table \ref{tab:Enrollment}. 

We do see that Doctor2Vec is able to outperform our two ablation, \framework-style models trained on the same smaller, full-data training dataset with less than 5000 trials. However, \framework's missing data mechanism unlocks an effective form of data augmentation which allows it to make use of the larger augmented missing-data training dataset. Accordingly, it is able to achieve the best enrollment performance in both metrics on this full-data test set, even though it was trained on a different type of data.

\begin{figure}
    \centering
    \includegraphics[scale=0.44]{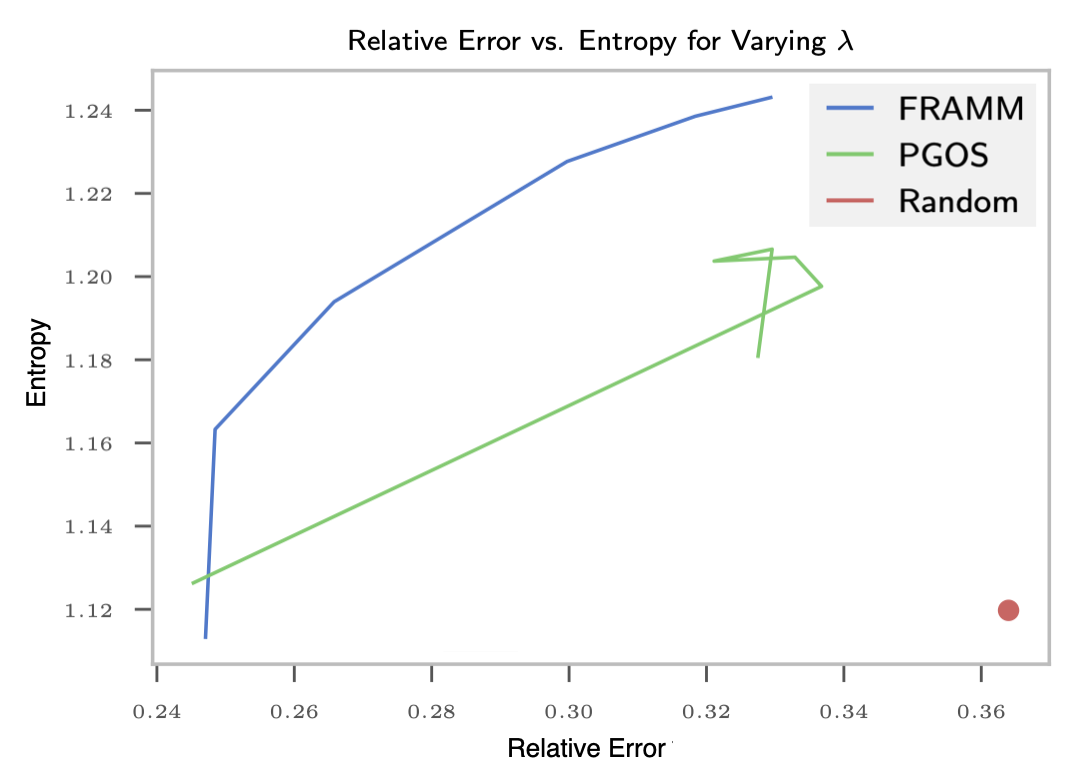}
    \caption{Relative error vs. entropy tradeoff curves for $\lambda$ equaling 0.5, 1, 2, 4, and 8 for $M = 10$, $K = 5$. Both \framework and the PGOS model are able to increase diversity at the expense of enrollment, but \framework makes much more efficient and tunable trade-offs than the PGOS baseline. It maintains much higher enrollment rates for a given level of diversity, achieving a roughly 5\% higher peak diversity value and providing up to 9\% higher levels of diversity for the same enrollment value.}
    \label{fig:Tradeoffs}

\end{figure}

\begin{figure}
    \centering
    \includegraphics[scale=0.35]{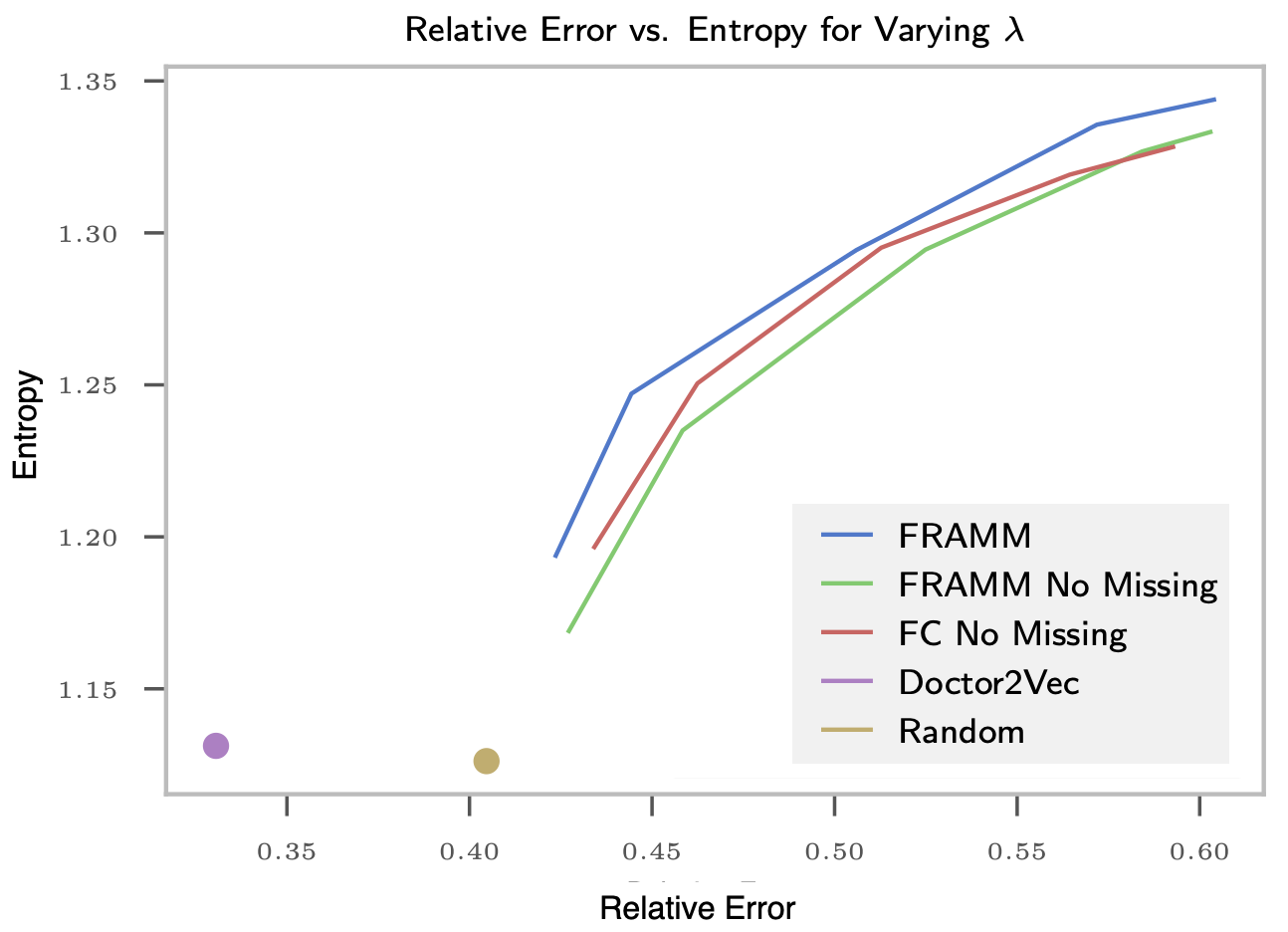}
    \caption{Relative error vs. entropy tradeoff curves comparing \framework to our two ablation models and the two enrollment baselines on the full-data test set. \framework leverages training with missing data to make more efficient trade-offs than either ablation baseline, and achieves more optimal combinations of enrollment and diversity than the Doctor2Vec and Random baselines which are constrained to a single point without any ability to increase diversity.}
    \label{fig:AugmentationTradeoffs}

\end{figure}

\subsection*{Q2. Making Efficient Trade-offs}
We then showcase our framework’s ability to make effective tradeoffs between enrollment and diversity by showing the trajectories of relative error vs. entropy for varying $\lambda$ values. 

We compare \framework to our PGOS and Random baselines in the the M=10, K=5 setting on the missing data test set in Figure \ref{fig:Tradeoffs}. Here we see that while both \framework and the PGOS model greatly outperform the Random baseline and are able to increase diversity at the expense of enrollment, \framework makes much more efficient and tunable trade-offs than the PGOS baseline. As such, it maintains much higher enrollment rates for a given level of diversity, achieving a roughly 5\% higher peak diversity value and providing up to 9\% higher levels of diversity for the same enrollment value. Furthermore, it offers the ability for much more granular tuning through different $\lambda$ values whereas the PGOS model is largely constrained to the same region once $\lambda$ is increased from 0. 

We also compare \framework to our two ablation models and the two enrollment baselines on the full-data test set in Figure \ref{fig:AugmentationTradeoffs}. There we see again that \framework is effectively leveraging training with missing data in making more efficient trade-offs than either ablation baseline. Finally, we see that it is able to achieve more optimal combinations of enrollment and diversity than the Doctor2Vec and Random baselines which are constrained to a single point without any ability to increase diversity.

\subsection*{Q3. Improving Diversity}
Finally, we examine the effect of our model on the enrolled populations of the studies themselves as compared to those selected by the enrollment-only Doctor2Vec model, presenting both the aggregate effect and the effect on a single randomly selected study about multiple sclerosis.

\begin{figure}
    \centering
    \includegraphics[scale=0.8]{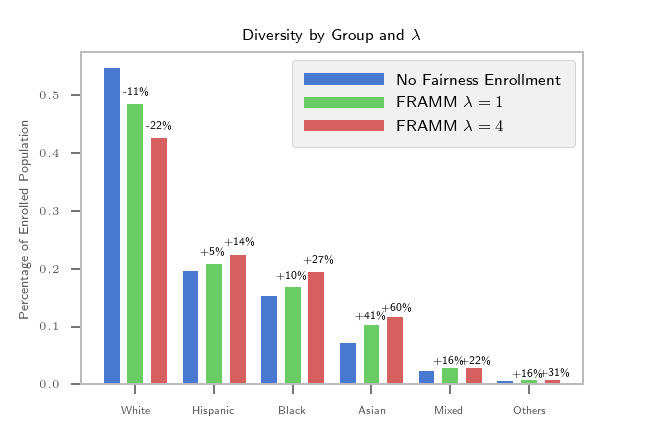}
    \caption{The aggregate effect of different $\lambda$ values as compared to enrollment of our Doctor2Vec baseline without consideration of fairness on the racial make-up of the average enrolled population. We see a dramatic decrease of white enrollment and a clear increase in underrepresented groups.}
    \label{fig:Aggregate}
\end{figure}

\paragraph{Aggregate Effect}
The aggregate effects are striking in terms of improving diversity. We see a big reduction of the enrollment of white population with a corresponding increase in each minority group, with the most significant increase for Black and Asian groups. The comparison of the mean percentages of each racial group across the trials in the test set for Doctor2Vec's chosen cohorts, the population enrolled by \framework with $\lambda = 1$, and the population enrolled by \framework with $\lambda = 4$ can be seen in Figure \ref{fig:Aggregate}.

\paragraph{Individual Effect}
This effect is mirrored in the case of a single, randomly selected trial for Relapsing Multiple Sclerosis (RMS). We present comprehensive results including all possible site options in Table \ref{tab:CaseStudyOptions}. In comparison to the sites chosen by Doctor2Vec, we see a concerted shift from sites with overwhelmingly white populations to those which enroll just as many but possess much more diversity. Specifically, \framework can be seen as extracting the optimal cohort in selecting each of the top 5 racially diverse sites. As a result, the racial distribution of sites changes from $[56.1, 15.8, 18.1, 6.5, 2.8, 0.7]$ to $[45.9, 15.5, 26.2,  7.0,  4.2, 0.9]$, increasing the entropy of the enrolled population from 1.240 to 1.362 while simultaneously enrolling more patients.

This performance appears even stronger when put in the context of other options for improving diversity. We already saw in the previous sections that our PGOS fairness baseline made less efficient trade-offs than \framework. However, if a trial decided to forgo a reinforcement learning setup altogether and not optimize for diversity but instead constrain fairness within the selections of an enrollment-only model, the enrollment to achieve that level of diversity would be incredibly low. Specifically in the case of this study, forcing an acceptable level of diversity (for example matching \framework's cohort distribution) on Doctor2Vec's 5 selected sites through enrollment caps would further reduce its enrollment by at least 18\% just to reduce the white population to the same proportion.
\begin{table}
\caption{Case study trial site selection}
\begin{tabular}{l|cccccc|c} \toprule
        \tiny Site Location & \tiny White & \tiny Hispanic & \tiny Black & \tiny Asian & \tiny Mixed & \tiny Others & \tiny Enrolled \\ \midrule
     \tiny Birmingham, AL &   67.7 &       3.8 &   25.5 &    1.7 &    1.2 &     0.1 &                 23.0 \\
 \tiny Wellesley Hills, MA &   80.4 &       4.2 &    2.0 &    9.6 &    3.2 &     0.5 &                 17.0 \\
         \tiny{\textbf{Tacoma, WA}} &   
         {\textbf{60.0}} &      {\textbf{10.9}} &   {\textbf{14.5}} &    {\textbf{5.2}} &    {\textbf{7.8}} &     {\textbf{1.6}} &                 {\textbf{16.0}} \\
      \tiny Ann Arbor, MI &   66.9 &       3.7 &    7.5 &   17.9 &    3.9 &     0.1 &                 16.0 \\
 \tiny\underline{\textbf{Fort Lauderdale, FL}} &   \underline{\textbf{29.5}} &      \underline{\textbf{25.2}} &   \underline{\textbf{39.4}} &    \underline{\textbf{4.0}} &    \underline{\textbf{1.8}} &     \underline{\textbf{0.1}} &                 \underline{\textbf{14.0}} \\
 \tiny\underline{\textbf{San Antonio, TX}} &   \underline{\textbf{53.4}} &      \underline{\textbf{30.8}} &    \underline{\textbf{5.2}} &    \underline{\textbf{8.3}} &    \underline{\textbf{1.6}} &     \underline{\textbf{0.6}} &                 \underline{\textbf{12.0}} \\
        \tiny Raleigh, NC &   73.3 &       6.2 &   10.6 &    6.5 &    2.8 &     0.5 &                 12.0 \\
       \tiny Kirkland, WA &   69.9 &       8.1 &    1.4 &   15.8 &    4.2 &     0.7 &                 12.0 \\
 \tiny\underline{\textbf{Oklahoma City, OK}} &   \underline{\textbf{42.0}} &       \underline{\textbf{3.0}} &   \underline{\textbf{36.1}} &   \underline{\textbf{10.6}} &    \underline{\textbf{6.4}} &     \underline{\textbf{1.9}} &                 \underline{\textbf{11.0}} \\
         \tiny \underline{Tucson, AZ} &   \underline{77.7} &      \underline{12.4} &    \underline{1.4} &    \underline{5.7} &    \underline{1.8} &     \underline{1.1} &                 \underline{11.0} \\
     \tiny  \underline{Knoxville, TN} &   \underline{88.2} &       \underline{2.4} &    \underline{2.3} &    \underline{4.1} &    \underline{2.6} &     \underline{0.4} &                 \underline{10.0} \\
     \tiny  Charlotte, NC &   70.2 &       2.9 &   19.5 &    3.8 &    2.7 &     0.9 &                  9.0 \\
      \tiny Asheville, NC &   76.6 &       9.6 &    8.0 &    1.8 &    3.2 &     0.9 &                  9.0 \\
          \tiny Greer, SC &   76.1 &       9.3 &    7.8 &    4.9 &    1.2 &     0.7 &                  9.0 \\
      \tiny{\textbf{Cleveland, OH}} &   {\textbf{40.9}} &      {\textbf{2.5}} &   {\textbf{44.7}} &    {\textbf{9.5}} &    {\textbf{2.3}} &     {\textbf{0.2}} &                  {\textbf{8.0}} \\
         \tiny Owosso, MI &   93.9 &       3.4 &    0.5 &    0.6 &    1.4 &     0.3 &                  8.0 \\
         \tiny Toledo, OH &   85.3 &       3.6 &    6.0 &    2.5 &    1.9 &     0.6 &                  8.0 \\
     \tiny Louisville, CO &   80.0 &       6.5 &    0.4 &   10.7 &    2.2 &     0.2 &                  8.0 \\
      \tiny Flossmoor, IL &   42.5 &       3.0 &   50.3 &    1.3 &    1.5 &     1.4 &                  7.0 \\
        \tiny Cullman, AL &   95.2 &       4.0 &    0.4 &    0.2 &    0.2 &     0.0 &                  7.0 \\ \bottomrule
\end{tabular}
\label{tab:CaseStudyOptions}
\leavevmode
\end{table}

\section{Conclusion}
We propose a deep reinforcement learning framework, named \framework, for fair trial site selection. Our method uses a missing data mechanism to account for the fact that different modalities of input features can be missing at different sites. It also uses reinforcement learning with a specifically designed reward function that simultaneously optimizes for both enrollment and fairness to account for the need to make efficient trade-offs between the two objectives.
We demonstrate strong performance in achieving state of the art enrollment levels and also the ability to make efficient and tunable trade-offs between enrollment and diversity. Finally, we show that when diversity is increased, we achieve much more fair site selection in enrolling far more underrepresented populations than other enrollment-only models.


\bibliographystyle{ACM-Reference-Format}
\bibliography{refs}


\begin{thebibliography}{24}


\ifx \showCODEN    \undefined \def \showCODEN     #1{\unskip}     \fi
\ifx \showDOI      \undefined \def \showDOI       #1{#1}\fi
\ifx \showISBNx    \undefined \def \showISBNx     #1{\unskip}     \fi
\ifx \showISBNxiii \undefined \def \showISBNxiii  #1{\unskip}     \fi
\ifx \showISSN     \undefined \def \showISSN      #1{\unskip}     \fi
\ifx \showLCCN     \undefined \def \showLCCN      #1{\unskip}     \fi
\ifx \shownote     \undefined \def \shownote      #1{#1}          \fi
\ifx \showarticletitle \undefined \def \showarticletitle #1{#1}   \fi
\ifx \showURL      \undefined \def \showURL       {\relax}        \fi
\providecommand\bibfield[2]{#2}
\providecommand\bibinfo[2]{#2}
\providecommand\natexlab[1]{#1}
\providecommand\showeprint[2][]{arXiv:#2}

\bibitem[\protect\citeauthoryear{??}{USC}{2018}]%
        {USC}
 \bibinfo{year}{2018}\natexlab{}.
\newblock \bibinfo{booktitle}{\emph{The Uniform System of Classification
  (USC)}}.
\newblock \bibinfo{type}{Report}. \bibinfo{institution}{Centers for Disease
  Control and Prevention}.
\newblock
\urldef\tempurl%
\url{https://www.cdc.gov/antibiotic-use/community/pdfs/Uniform-System-of-Classification-2018-p.pdf}
\showURL{%
\tempurl}


\bibitem[\protect\citeauthoryear{Biswal, Xiao, Glass, Milkovits, and
  Sun}{Biswal et~al\mbox{.}}{2020}]%
        {biswal2020doctor2vec}
\bibfield{author}{\bibinfo{person}{Siddharth Biswal}, \bibinfo{person}{Cao
  Xiao}, \bibinfo{person}{Lucas~M Glass}, \bibinfo{person}{Elizabeth
  Milkovits}, {and} \bibinfo{person}{Jimeng Sun}.}
  \bibinfo{year}{2020}\natexlab{}.
\newblock \showarticletitle{Doctor2vec: Dynamic doctor representation learning
  for clinical trial recruitment}. In \bibinfo{booktitle}{\emph{Proceedings of
  the AAAI Conference on Artificial Intelligence}}, Vol.~\bibinfo{volume}{34}.
  \bibinfo{pages}{557--564}.
\newblock


\bibitem[\protect\citeauthoryear{Fu, Huang, Xiao, Glass, and Sun}{Fu
  et~al\mbox{.}}{2022}]%
        {fu2022hint}
\bibfield{author}{\bibinfo{person}{Tianfan Fu}, \bibinfo{person}{Kexin Huang},
  \bibinfo{person}{Cao Xiao}, \bibinfo{person}{Lucas~M Glass}, {and}
  \bibinfo{person}{Jimeng Sun}.} \bibinfo{year}{2022}\natexlab{}.
\newblock \showarticletitle{HINT: Hierarchical interaction network for
  clinical-trial-outcome predictions}.
\newblock \bibinfo{journal}{\emph{Patterns}} \bibinfo{volume}{3},
  \bibinfo{number}{4} (\bibinfo{year}{2022}), \bibinfo{pages}{100445}.
\newblock


\bibitem[\protect\citeauthoryear{Gao, Xiao, Glass, and Sun}{Gao
  et~al\mbox{.}}{2020}]%
        {gao2020compose}
\bibfield{author}{\bibinfo{person}{Junyi Gao}, \bibinfo{person}{Cao Xiao},
  \bibinfo{person}{Lucas~M Glass}, {and} \bibinfo{person}{Jimeng Sun}.}
  \bibinfo{year}{2020}\natexlab{}.
\newblock \showarticletitle{COMPOSE: Cross-Modal Pseudo-Siamese Network for
  Patient Trial Matching}. In \bibinfo{booktitle}{\emph{Proceedings of the 26th
  ACM SIGKDD International Conference on Knowledge Discovery \& Data Mining}}.
  \bibinfo{pages}{803--812}.
\newblock


\bibitem[\protect\citeauthoryear{Gligorijevic, Gligorijevic, Pavlovski,
  Milkovits, Glass, Grier, Vankireddy, and Obradovic}{Gligorijevic
  et~al\mbox{.}}{2019}]%
        {gligorijevic2019optimizing}
\bibfield{author}{\bibinfo{person}{Jelena Gligorijevic},
  \bibinfo{person}{Djordje Gligorijevic}, \bibinfo{person}{Martin Pavlovski},
  \bibinfo{person}{Elizabeth Milkovits}, \bibinfo{person}{Lucas Glass},
  \bibinfo{person}{Kevin Grier}, \bibinfo{person}{Praveen Vankireddy}, {and}
  \bibinfo{person}{Zoran Obradovic}.} \bibinfo{year}{2019}\natexlab{}.
\newblock \showarticletitle{Optimizing clinical trials recruitment via deep
  learning}.
\newblock \bibinfo{journal}{\emph{Journal of the American Medical Informatics
  Association}} \bibinfo{volume}{26}, \bibinfo{number}{11}
  (\bibinfo{year}{2019}), \bibinfo{pages}{1195--1202}.
\newblock


\bibitem[\protect\citeauthoryear{Gray, Nolan, Gregory, and Joseph}{Gray
  et~al\mbox{.}}{2021}]%
        {gray2021diversity}
\bibfield{author}{\bibinfo{person}{Darrell~M Gray}, \bibinfo{person}{Timiya~S
  Nolan}, \bibinfo{person}{John Gregory}, {and} \bibinfo{person}{Joshua~J
  Joseph}.} \bibinfo{year}{2021}\natexlab{}.
\newblock \showarticletitle{Diversity in clinical trials: an opportunity and
  imperative for community engagement}.
\newblock \bibinfo{journal}{\emph{The Lancet Gastroenterology \& Hepatology}}
  \bibinfo{volume}{6}, \bibinfo{number}{8} (\bibinfo{year}{2021}),
  \bibinfo{pages}{605--607}.
\newblock


\bibitem[\protect\citeauthoryear{Hughson, Woodward-Kron, Parker, Hajek, Bresin,
  Knoch, Phan, and Story}{Hughson et~al\mbox{.}}{2016}]%
        {hughson2016review}
\bibfield{author}{\bibinfo{person}{Jo-anne Hughson}, \bibinfo{person}{Robyn
  Woodward-Kron}, \bibinfo{person}{Anna Parker}, \bibinfo{person}{John Hajek},
  \bibinfo{person}{Agnese Bresin}, \bibinfo{person}{Ute Knoch},
  \bibinfo{person}{Tuong Phan}, {and} \bibinfo{person}{David Story}.}
  \bibinfo{year}{2016}\natexlab{}.
\newblock \showarticletitle{A review of approaches to improve participation of
  culturally and linguistically diverse populations in clinical trials}.
\newblock \bibinfo{journal}{\emph{Trials}} \bibinfo{volume}{17},
  \bibinfo{number}{1} (\bibinfo{year}{2016}), \bibinfo{pages}{1--10}.
\newblock


\bibitem[\protect\citeauthoryear{Hwang and Brawley}{Hwang and Brawley}{2022}]%
        {hwang2022new}
\bibfield{author}{\bibinfo{person}{Thomas~J Hwang} {and}
  \bibinfo{person}{Otis~W Brawley}.} \bibinfo{year}{2022}\natexlab{}.
\newblock \showarticletitle{New federal incentives for diversity in clinical
  trials}.
\newblock \bibinfo{journal}{\emph{New England Journal of Medicine}}
  \bibinfo{volume}{387}, \bibinfo{number}{15} (\bibinfo{year}{2022}),
  \bibinfo{pages}{1347--1349}.
\newblock


\bibitem[\protect\citeauthoryear{Knepper and McLeod}{Knepper and
  McLeod}{2018}]%
        {knepper2018will}
\bibfield{author}{\bibinfo{person}{Todd~C Knepper} {and}
  \bibinfo{person}{Howard~L McLeod}.} \bibinfo{year}{2018}\natexlab{}.
\newblock \bibinfo{title}{When will clinical trials finally reflect diversity?}
\newblock
\newblock


\bibitem[\protect\citeauthoryear{Lau, Adler, and Sj{\"o}lund}{Lau
  et~al\mbox{.}}{2019}]%
        {lau2019unified}
\bibfield{author}{\bibinfo{person}{Kenneth Lau}, \bibinfo{person}{Jonas Adler},
  {and} \bibinfo{person}{Jens Sj{\"o}lund}.} \bibinfo{year}{2019}\natexlab{}.
\newblock \showarticletitle{A unified representation network for segmentation
  with missing modalities}.
\newblock \bibinfo{journal}{\emph{arXiv preprint arXiv:1908.06683}}
  (\bibinfo{year}{2019}).
\newblock


\bibitem[\protect\citeauthoryear{Liu, Carter, and Gifford}{Liu
  et~al\mbox{.}}{2021a}]%
        {Liu2021-kc}
\bibfield{author}{\bibinfo{person}{Ge Liu}, \bibinfo{person}{Brandon Carter},
  {and} \bibinfo{person}{David~K Gifford}.} \bibinfo{year}{2021}\natexlab{a}.
\newblock \showarticletitle{Predicted Cellular Immunity Population Coverage
  Gaps for {SARS-CoV-2} Subunit Vaccines and Their Augmentation by Compact
  Peptide Sets}.
\newblock \bibinfo{journal}{\emph{Cell Syst}} \bibinfo{volume}{12},
  \bibinfo{number}{1} (\bibinfo{date}{Jan.} \bibinfo{year}{2021}),
  \bibinfo{pages}{102--107.e4}.
\newblock


\bibitem[\protect\citeauthoryear{Liu, Rizzo, Whipple, Pal, Pineda, Lu, Arnieri,
  Lu, Capra, Copping, et~al\mbox{.}}{Liu et~al\mbox{.}}{2021b}]%
        {liu2021evaluating}
\bibfield{author}{\bibinfo{person}{Ruishan Liu}, \bibinfo{person}{Shemra
  Rizzo}, \bibinfo{person}{Samuel Whipple}, \bibinfo{person}{Navdeep Pal},
  \bibinfo{person}{Arturo~Lopez Pineda}, \bibinfo{person}{Michael Lu},
  \bibinfo{person}{Brandon Arnieri}, \bibinfo{person}{Ying Lu},
  \bibinfo{person}{William Capra}, \bibinfo{person}{Ryan Copping},
  {et~al\mbox{.}}} \bibinfo{year}{2021}\natexlab{b}.
\newblock \showarticletitle{Evaluating eligibility criteria of oncology trials
  using real-world data and AI}.
\newblock \bibinfo{journal}{\emph{Nature}} \bibinfo{volume}{592},
  \bibinfo{number}{7855} (\bibinfo{year}{2021}), \bibinfo{pages}{629--633}.
\newblock


\bibitem[\protect\citeauthoryear{Ma, Xu, Huang, and Zhang}{Ma
  et~al\mbox{.}}{2021b}]%
        {ma2021maximum}
\bibfield{author}{\bibinfo{person}{Fei Ma}, \bibinfo{person}{Xiangxiang Xu},
  \bibinfo{person}{Shao-Lun Huang}, {and} \bibinfo{person}{Lin Zhang}.}
  \bibinfo{year}{2021}\natexlab{b}.
\newblock \showarticletitle{Maximum Likelihood Estimation for Multimodal
  Learning with Missing Modality}.
\newblock \bibinfo{journal}{\emph{arXiv preprint arXiv:2108.10513}}
  (\bibinfo{year}{2021}).
\newblock


\bibitem[\protect\citeauthoryear{Ma, Ren, Zhao, Tulyakov, Wu, and Peng}{Ma
  et~al\mbox{.}}{2021a}]%
        {ma2021smil}
\bibfield{author}{\bibinfo{person}{Mengmeng Ma}, \bibinfo{person}{Jian Ren},
  \bibinfo{person}{Long Zhao}, \bibinfo{person}{Sergey Tulyakov},
  \bibinfo{person}{Cathy Wu}, {and} \bibinfo{person}{Xi Peng}.}
  \bibinfo{year}{2021}\natexlab{a}.
\newblock \showarticletitle{Smil: Multimodal learning with severely missing
  modality}. In \bibinfo{booktitle}{\emph{Proceedings of the AAAI Conference on
  Artificial Intelligence}}, Vol.~\bibinfo{volume}{35}.
  \bibinfo{pages}{2302--2310}.
\newblock


\bibitem[\protect\citeauthoryear{Nephew}{Nephew}{2021}]%
        {nephew2021accountability}
\bibfield{author}{\bibinfo{person}{Lauren~D Nephew}.}
  \bibinfo{year}{2021}\natexlab{}.
\newblock \showarticletitle{Accountability in clinical trial diversity: The
  buck stops where?}
\newblock \bibinfo{journal}{\emph{EClinicalMedicine}}  \bibinfo{volume}{36}
  (\bibinfo{year}{2021}).
\newblock


\bibitem[\protect\citeauthoryear{Parthasarathy and Sundaram}{Parthasarathy and
  Sundaram}{2020}]%
        {parthasarathy2020training}
\bibfield{author}{\bibinfo{person}{Srinivas Parthasarathy} {and}
  \bibinfo{person}{Shiva Sundaram}.} \bibinfo{year}{2020}\natexlab{}.
\newblock \showarticletitle{Training strategies to handle missing modalities
  for audio-visual expression recognition}. In
  \bibinfo{booktitle}{\emph{Companion Publication of the 2020 International
  Conference on Multimodal Interaction}}. \bibinfo{pages}{400--404}.
\newblock


\bibitem[\protect\citeauthoryear{Paszke, Gross, Massa, Lerer, Bradbury, Chanan,
  Killeen, Lin, Gimelshein, Antiga, Desmaison, Kopf, Yang, DeVito, Raison,
  Tejani, Chilamkurthy, Steiner, Fang, Bai, and Chintala}{Paszke
  et~al\mbox{.}}{2019}]%
        {PyTorch}
\bibfield{author}{\bibinfo{person}{Adam Paszke}, \bibinfo{person}{Sam Gross},
  \bibinfo{person}{Francisco Massa}, \bibinfo{person}{Adam Lerer},
  \bibinfo{person}{James Bradbury}, \bibinfo{person}{Gregory Chanan},
  \bibinfo{person}{Trevor Killeen}, \bibinfo{person}{Zeming Lin},
  \bibinfo{person}{Natalia Gimelshein}, \bibinfo{person}{Luca Antiga},
  \bibinfo{person}{Alban Desmaison}, \bibinfo{person}{Andreas Kopf},
  \bibinfo{person}{Edward Yang}, \bibinfo{person}{Zachary DeVito},
  \bibinfo{person}{Martin Raison}, \bibinfo{person}{Alykhan Tejani},
  \bibinfo{person}{Sasank Chilamkurthy}, \bibinfo{person}{Benoit Steiner},
  \bibinfo{person}{Lu Fang}, \bibinfo{person}{Junjie Bai}, {and}
  \bibinfo{person}{Soumith Chintala}.} \bibinfo{year}{2019}\natexlab{}.
\newblock \showarticletitle{PyTorch: An Imperative Style, High-Performance Deep
  Learning Library}.
\newblock In \bibinfo{booktitle}{\emph{Advances in Neural Information
  Processing Systems 32}}. \bibinfo{publisher}{Curran Associates, Inc.},
  \bibinfo{pages}{8024--8035}.
\newblock
\urldef\tempurl%
\url{http://papers.neurips.cc/paper/9015-pytorch-an-imperative-style-high-performance-deep-learning-library.pdf}
\showURL{%
\tempurl}


\bibitem[\protect\citeauthoryear{Sharma and Palaniappan}{Sharma and
  Palaniappan}{2021}]%
        {sharma2021improving}
\bibfield{author}{\bibinfo{person}{Ashwarya Sharma} {and}
  \bibinfo{person}{Latha Palaniappan}.} \bibinfo{year}{2021}\natexlab{}.
\newblock \showarticletitle{Improving diversity in medical research}.
\newblock \bibinfo{journal}{\emph{Nature Reviews Disease Primers}}
  \bibinfo{volume}{7}, \bibinfo{number}{1} (\bibinfo{year}{2021}),
  \bibinfo{pages}{1--2}.
\newblock


\bibitem[\protect\citeauthoryear{Singh and Joachims}{Singh and
  Joachims}{2019}]%
        {singh2019policy}
\bibfield{author}{\bibinfo{person}{Ashudeep Singh} {and}
  \bibinfo{person}{Thorsten Joachims}.} \bibinfo{year}{2019}\natexlab{}.
\newblock \showarticletitle{Policy learning for fairness in ranking}.
\newblock \bibinfo{journal}{\emph{arXiv preprint arXiv:1902.04056}}
  (\bibinfo{year}{2019}).
\newblock


\bibitem[\protect\citeauthoryear{Tran, Liu, Zhou, and Jin}{Tran
  et~al\mbox{.}}{2017}]%
        {Tran_2017_CVPR}
\bibfield{author}{\bibinfo{person}{Luan Tran}, \bibinfo{person}{Xiaoming Liu},
  \bibinfo{person}{Jiayu Zhou}, {and} \bibinfo{person}{Rong Jin}.}
  \bibinfo{year}{2017}\natexlab{}.
\newblock \showarticletitle{Missing Modalities Imputation via Cascaded Residual
  Autoencoder}. In \bibinfo{booktitle}{\emph{Proceedings of the IEEE Conference
  on Computer Vision and Pattern Recognition (CVPR)}}.
\newblock


\bibitem[\protect\citeauthoryear{Wang and Sun}{Wang and Sun}{2022}]%
        {wang2022trial2vec}
\bibfield{author}{\bibinfo{person}{Zifeng Wang} {and} \bibinfo{person}{Jimeng
  Sun}.} \bibinfo{year}{2022}\natexlab{}.
\newblock \showarticletitle{Trial2Vec: Zero-Shot Clinical Trial Document
  Similarity Search using Self-Supervision}.
\newblock \bibinfo{journal}{\emph{arXiv preprint arXiv:2206.14719}}
  (\bibinfo{year}{2022}).
\newblock


\bibitem[\protect\citeauthoryear{Williams}{Williams}{1992}]%
        {williams1992simple}
\bibfield{author}{\bibinfo{person}{Ronald~J Williams}.}
  \bibinfo{year}{1992}\natexlab{}.
\newblock \showarticletitle{Simple statistical gradient-following algorithms
  for connectionist reinforcement learning}.
\newblock \bibinfo{journal}{\emph{Machine learning}} \bibinfo{volume}{8},
  \bibinfo{number}{3} (\bibinfo{year}{1992}), \bibinfo{pages}{229--256}.
\newblock


\bibitem[\protect\citeauthoryear{Yates, Byrne, Donahue, McCarty, and
  Mathews}{Yates et~al\mbox{.}}{2020}]%
        {yates2020representation}
\bibfield{author}{\bibinfo{person}{Isabelle Yates}, \bibinfo{person}{Jennifer
  Byrne}, \bibinfo{person}{S Donahue}, \bibinfo{person}{Linda McCarty}, {and}
  \bibinfo{person}{Allison Mathews}.} \bibinfo{year}{2020}\natexlab{}.
\newblock \showarticletitle{Representation in clinical trials: A review on
  reaching underrepresented populations in research}.
\newblock \bibinfo{journal}{\emph{Clinical Researcher}} \bibinfo{volume}{34},
  \bibinfo{number}{7} (\bibinfo{year}{2020}).
\newblock


\bibitem[\protect\citeauthoryear{Zhang, Xiao, Glass, and Sun}{Zhang
  et~al\mbox{.}}{2020}]%
        {zhang2020deepenroll}
\bibfield{author}{\bibinfo{person}{Xingyao Zhang}, \bibinfo{person}{Cao Xiao},
  \bibinfo{person}{Lucas~M Glass}, {and} \bibinfo{person}{Jimeng Sun}.}
  \bibinfo{year}{2020}\natexlab{}.
\newblock \showarticletitle{Deepenroll: Patient-trial matching with deep
  embedding and entailment prediction}. In
  \bibinfo{booktitle}{\emph{Proceedings of The Web Conference 2020}}.
  \bibinfo{pages}{1029--1037}.
\newblock


\end{thebibliography}

\appendix

\section{Experimental Details}
Here we provide some concrete details about the data preparation used to create our real-world dataset and then the setup and hyperparameters which went into model training and validation.

\subsection{Input Feature Specifics}
We outline the features that compose the representation vectors and matrices for the trial and site representations in our main paper, but we provide additional details regarding their construction and dimensionality here. $n_t$ and $n_s$, representing the dimensionality of the trial feature vector and the static site feature vector respectively, are equal to 1827 and 669. The one-hot vectors representing a site's diagnosis history have dimensionality $n_d = 260$ for ICD-10 code categories defined by the first letter and second number of a given code, representing its higher level category in the wider ICD-10 ontology. Similarly, the prescription history one-hot vectors have dimensionlaity $n_r = 100$ for the first two digits in each code within the USC ontology \cite{USC} .

\subsection{Data preparation}
For our experiments, we use real-world clinical trials and claims data to train and evaluate our algorithm. The clinical trial database contains 33,323 sites matched with 4,392 trials. We first build the site pool by for each site, creating the static features, matching the 500 most recent diagnoses and prescriptions, and setting the enrollment histories for the 50 most recent trials at any given point of time. We then create a separate dataset for each values of $M$ that we use. We match each trial to $M$ sites, using the top $M$ sites (determined by enrollment) in the database if there are enough and otherwise completing the set of $M$ by randomly selecting sites from the overall pool and assigning an enrollment of $0$ for the trial. Finally, we add in the missing data aspect by creating 10 versions of each trial, and for each site in each trial randomly creating a mask which dictates whether a given site modality is present for that data point, where each modality has an 80\% chance of being present, and we stipulate that at least one modality must be present for each site. After this augmentation, we are left with our final dataset for the given $M$ value.

\subsection{Model training and validation}
We then split this dataset into training, validation, and test datasets. Each split is random, with 20\% of overall dataset being reserved for use as a test set, and 10\% of the remaining training set serving as our validation set. Using these datasets, we train our models within the PyTorch framework \cite{PyTorch} for 35 epochs at a 0.00001 learning rate and using the Adam optimizer. We save the model which best performs on the validation set as determined by our loss function and evaluate it using the test set.

\begin{table}[]
\centering
\caption{Synthetic Enrollment-Only Performance}
\begin{tabular}{l|cc}
\toprule
&Relative Error &nDCG \\ \midrule
Random	 &0.227 $\pm$ 0.003	 &0.707 $\pm$ 0.003\\
FRAMM	 &0.062 $\pm$ 0.001	 &0.922 $\pm$ 0.002\\\bottomrule
\end{tabular}
\label{fig:EnrollmentStats}
\end{table}

\section{Synthetic Dataset}
We now outline the creation of and results on the synthetic dataset which was built to offer reproducibility of our work. All of our code for these experiments can be found at  \url{https://anonymous.4open.science/r/FRAMM-B4EB/}.

\begin{figure*}
\centering
\begin{subfigure}{0.5\textwidth}
    \captionsetup{skip=0pt}
    \centering
    \includegraphics[scale=0.66]{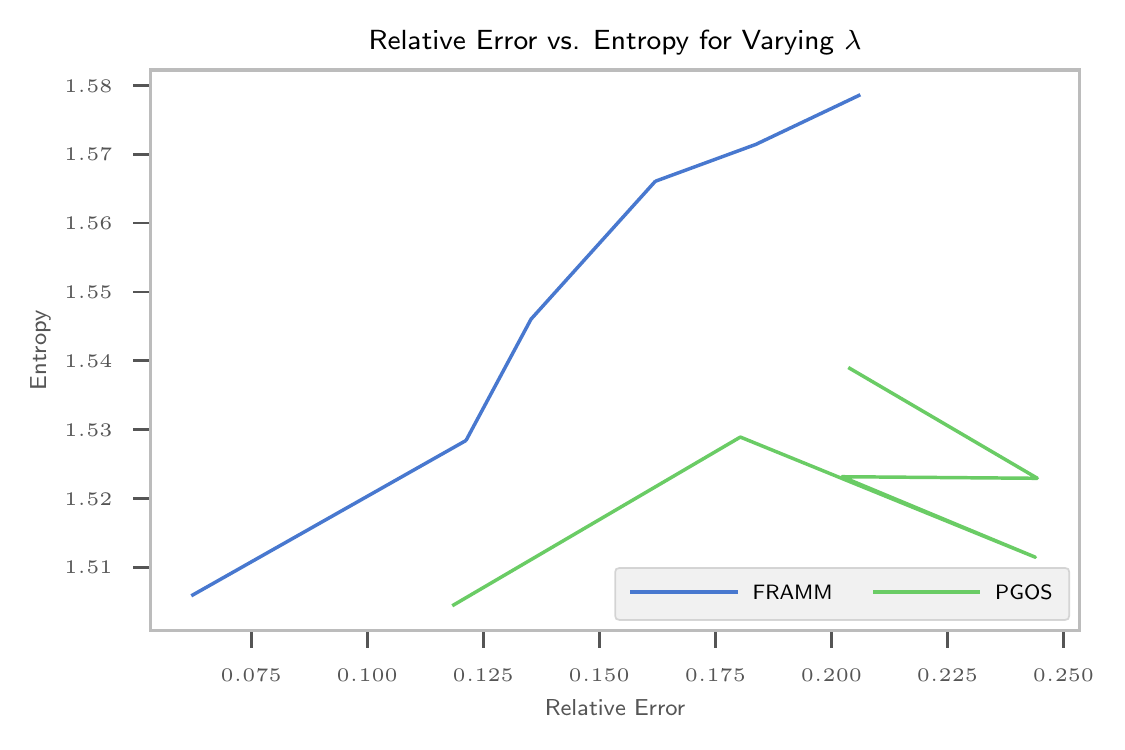}
    \caption[width=0.33\textwidth]{Missing Data Test Set}
    \label{fig:SuppTradeoffs}
\end{subfigure}%
\begin{subfigure}{0.5\textwidth}
    \captionsetup{skip=0pt}
    \centering
    \includegraphics[scale=0.66]{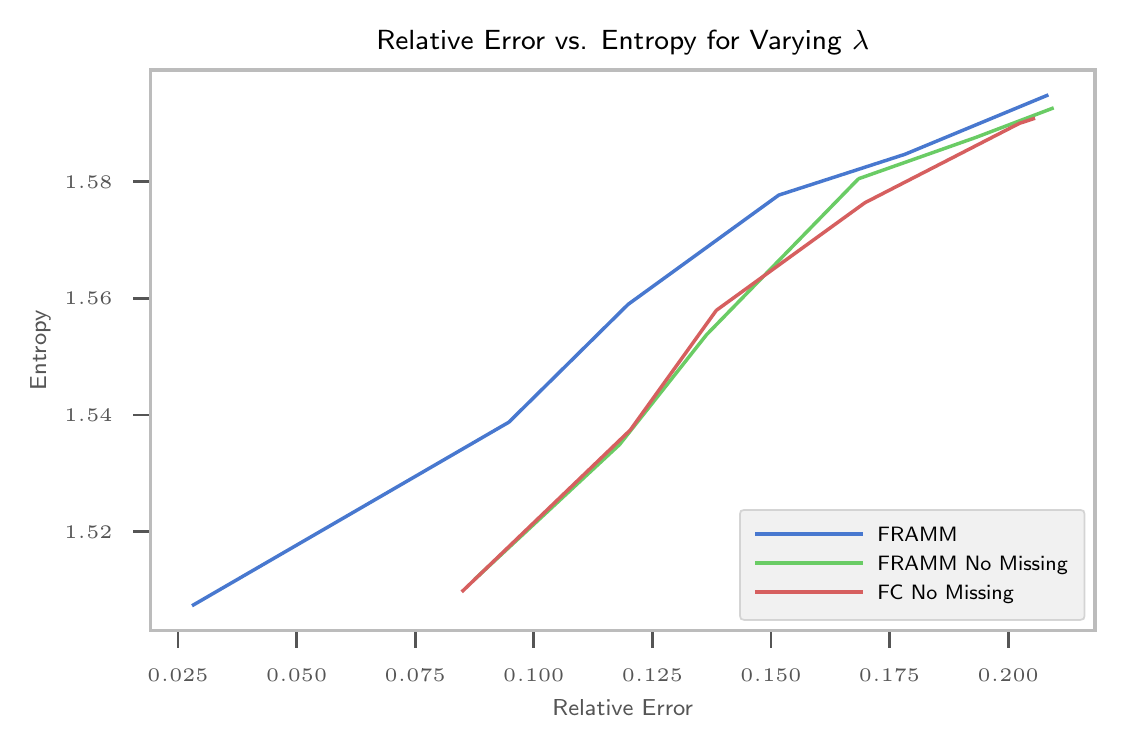}
    \caption[width=0.33\textwidth]{Full Data Test Set}
    \label{fig:SuppAugmentationTradeoffs}
\end{subfigure}
\captionsetup{skip=3pt}
\caption{Visualizations of relative error vs. entropy trade-offs for $M = 20$, $K = 10$, and $\lambda$ equaling 0, 0.5, 1, 2, 4, and 8 in each of our test settings on the synthetic dataset. Figure (a) compares \framework and the PGOS baseline on the core missing data test set where \framework is able to make much more efficient and tunable trade-offs than the PGOS baseline. Figure (b) shows the data augmentation experiment. While it was not trained on the complete (non-missing) data each model was evaluated on, we see a clear improvement via the \framework framework trained on a larger missing data dataset.}
\label{fig:AllTradeoffs}
\end{figure*}

\subsection{Dataset Creation}
The trial data itself is not proprietary as it was scraped from the publicly available \url{clinicaltrials.gov} website. So, we begin with the true trial representations. We then build a pool of 30,000 sites as follows. We first randomly build each of the static modalities features except for the primary specialty and racial distribution. The primary specialty is sampled from the percentages of each specialty within the true dataset, and the racial demographic distribution values are sampled from normal distributions centered roughly at the aggregate distributions from the true dataset and normalized. All of the features are then concatenated to form the static features modality for each synthetic site. We then build the diagnosis and prescription history modalities by sampling bigram probabilities for each of the 500 values from the true dataset values (with the initial value conditioning the first bigram being the primary specialty).

Now that we have our trials and our pool of sites (they each start with no enrollment history), we move on to simulating trials in order to generate the enrollment history and full synthetic dataset. We first train a model on the real dataset to predict the enrollment of a site for a given trial. This model acts as our data labeler. We then randomly order the trial representations before proceeding through them sequentially to generate the data. For each trial, we randomly select 20 sites from the pool and pass them in (with their current enrollment history) to the labeler. The labeler's outputs are used as the enrollment labels and combine with the site modalities and trial representation to build a single datapoint. We update each site's enrollment history with the trial representation and enrollment label and continue to the next trial.

At the end, we augment the produced dataset with missing modality masks in the same way as we did the real dataset and save the result as the synthetic dataset. The code to build this dataset (and necessary artifacts such as the labeler model and bigram probabilities) are all available within the GitHub repository we provide at \url{https://anonymous.4open.science/r/FRAMM-B4EB/}.

\subsection{Synthetic Results}
We now present our results on the synthetic dataset. We train each of our compared models on the $M=20$, $K=10$ setting for each of our $\lambda$ values (0, 0.5, 1, 2, 4, and 8) in order to generate a full set of results including for the data augmentation and serially missing modality experiments. The results of these synthetic data experiments qualitatively mirror those from the real-world dataset and are outlined below.

\subsubsection{Enrolling Large Patient Populations}
The enrollment-only results for $\lambda = 0$ on the synthetic dataset once again demonstrate a huge level of improvement over the random enrollment baseline. \framework here reduces relative error up to 74\% on the missing data test set. The full results of this experiment can be seen in Table \ref{fig:EnrollmentStats}.

\subsubsection{Balancing Enrollment and Fairness Trade-off}
The ability of different models to make effective trade-offs between enrollment and diversity on the synthetic dataset via varying $\lambda$ values is then shown in Figure \ref{fig:SuppTradeoffs}. \framework is generally able to make much more efficient and tunable trade-offs between our two objectives than the PGOS baseline. Furthermore, we show that \framework unlocks an effective version of data augmentation on the synthetic dataset. We display the trade-off results in Figure \ref{fig:SuppAugmentationTradeoffs} where we see that \framework easily outperforms the two variants trained only on the smaller, complete dataset. We specifically note that the \framework model trained on missing data is especially effective in more enrollment-based settings, outperforming the baselines for all $\lambda$ but reducing relative error up to 84\% when $\lambda = 0$.


\end{document}